\documentclass{article}

\PassOptionsToPackage{sort, numbers, square}{natbib}

\usepackage[preprint]{nips_2018}

\usepackage[utf8]{inputenc} 
\usepackage[T1]{fontenc}    
\usepackage{hyperref}       
\usepackage{url}            
\usepackage{booktabs}       
\usepackage{amsfonts}       
\usepackage{nicefrac}       
\usepackage{amsmath}
\usepackage[pdftex]{graphicx}

\title{Deep Learning Approximation: Zero-Shot Neural Network Speedup}

\author{
  Michele Pratusevich \\
  Amazon.com \\
  \texttt{pratusev@amazon.com}
}

\begin{document}

\maketitle

\begin{abstract}
  Neural networks offer high-accuracy solutions to a range of problems, but are costly to run in production systems because of computational and memory requirements during a forward pass. Given a trained network, we propose a techique called Deep Learning Approximation to build a faster network in a tiny fraction of the time required for training by only manipulating the network structure and coefficients without requiring re-training or access to the training data. Speedup is achieved by by applying a sequential series of independent optimizations that reduce the floating-point operations (FLOPs) required to perform a forward pass. First, lossless optimizations are applied, followed by lossy approximations using singular value decomposition (SVD) and low-rank matrix decomposition. The optimal approximation is chosen by weighing the relative accuracy loss and FLOP reduction according to a single parameter specified by the user. On PASCAL VOC 2007 with the YOLO network, we show an end-to-end 2x speedup in a network forward pass with a $5$\% drop in mAP that can be re-gained by finetuning.
\end{abstract}

\section{Introduction}

Deep Learning Approximation (DLA) speeds up the runtime of neural networks, which is especially relevant for pay-per-compute or limited-compute embedded environments. It decouples the task of speeding up neural networks from the task of making them more accurate. It enables tuning off-the-shelf networks for runtime, to complement fine-tuning for accuracy.

At deploy time, the dollar cost of a production pipeline using a neural network is directly proportional to the time required to execute a forward pass of the neural network. This is in contrast to training, which generally happens only once and offline. In turn, for large networks time is proportional to the number of floating-point operations (FLOPs) required in a forward pass. For example, in the cloud, instances are paid for by the hour, so a pipeline running a ResNet-50 model ($8 \times 10^9$ FLOPs) vs. a ResNet-152 model ($23 \times 10^9$ FLOPs) running $3$ times slower and therefore costs $3$x more. This extra dollar cost comes at minimal benefit ($22.9$\% vs $21.4\%$ error on ImageNet \citep{He2015}). Running faster models means a higher throughput, which means better hardware utilization, both in cloud and embedded environments. For embedded environments, faster and smaller models mean lower power, cooling, and CPU speed requirements.

DLA has several other important properties. For networks used as a black-box or where training data is confidential, re-training or fine-tuning the network is not an option or is extremely challenging. In these cases, DLA is a method for speeding up the network forward pass without needing access to the training data. If you can finetune, finetuning can recover accuracy lost during the DLA process.

In DLA, first apply lossless optimizations to reduce FLOP count and runtime, which is always an optimal optimization step. Then, we propose a method for automatically selecting an appropriate approximation to apply to each layer in a network based on a single parameter, which represents whether accuracy or speedup is more important. The approximation methods are all based on singular value decompositions (SVD) applied directly to the weight tensors.

On a benchmark set of standard computer vision tasks and neural network architectures, we show between a 1.5x and 2x speedup without additional re-training and minimal accuracy loss. In each case, funetuning after applying DLA recovers lost accuracy.

\section{Related Work}

\paragraph{Better hardware and software} To make a large, slow neural network run faster, you can deploy the net to faster hardware (FPGAs, better GPUs) using better software (such as MXNet \citep{DBLP:journals/corr/ChenLLLWWXXZZ15} and Caffe con Troll \citep{Hadjis:2015:CCT:2799562.2799641}). DLA offers a third option by operating directly on neural network models and can be applied to models targeting all hardware and software.

\paragraph{Training smaller models} Factorization and low-rank approximation approaches such as that of \citet{jaderberg14speeding} and factorized CNNs \citep{factorized_conv_nets} apply FLOP-reduction methods on convolution weights, but to inform neural network architecture decisions before training. Follow-up work on the spatial bottleneck architecture \citep{factorized_conv_nets}, SqueezeNet \cite{SqueezeNet}, and MobileNet \citep{DBLP:journals/corr/HowardZCKWWAA17} also focus on training architectures from the start that are faster or have fewer FLOPs. DLA can be applied after these models have been trained to achieve further speedup.

\paragraph{Training-in-the-loop methods} \citet{park_pruning} explore iteratively pruning and re-training individual convolutional filters to regain accuracy. Incremental quantization from \citep{zhou2017} adds quantization in the training loop as well, decreasing both memory and runtime. Similarly, \citet{wen_learning_2016} sparsifies convolutions during training time to take advantage of sparse matrix multiplication methods at deploy time. All these methods require access to the training data and are time-consuming. Whereas DLA can be applied on top of a network trained with any of these methods.

\paragraph{Model compression} Model compression makes model definitions smaller, especially when models need to be transferred in a limited-bandwidth or limited-memory environment \cite{Cheng2017ASO}. \citet{han2015deep_compression} combine iterative convolution pruning and Huffman coding to achieve high compression, but require extensive re-training to achieve accuracy parity. Binarized Neural Networks \citep{courbariaux_binarized_2016} and XNOR-Net \citep{rastegariECCV16} take compression even further, using only $1$ bit per weight value, but require access to the original training data to achieve high performance. DLA is complementary and can be applied after these methods have been used.

\paragraph{Data-agnostic approaches} \citet{Denton:2014:ELS:2968826.2968968} use SVD and factorization approaches to approximate convolutional and fully-connected layers, but also rely on re-training the network after an approximation has been applied to a single layer. Speedup is measured per-layer rather than end-to-end forward pass runtime, establishing a relationship between FLOPs and runtime for convolutional layers. DLA extends this method by applying approximations holistically to all layers in a network together, without requiring re-training in between applying approximations to individual layers. Additionally, the runtime decrease is measured end-to-end on a forward pass rather than on a per-layer basis.

\section{Runtime and FLOPs}

According to the roofline computation model \citep{roofline}, neural network runtime is dominated by FLOP count and memory movement to and from a specialized processor like a GPU. Convolutional, deconvolutional, and fully-connected layers (called \textbf{representation layers} throughout this paper) have a high arithmetic intensity, requiring many FLOPs to compute one unit of output. Therefore, these layers will operate in the FLOP-dominated regime from the roofline model rather than the memory-limited regime. CPUs and GPUs have a fixed processor throughput, so decreasing the number of FLOPs needed to perform an operation will decrease the amount of time spent in this bottleneck during a forward pass. DLA replaces FLOP-heavy representation layer operations with comparable operations that have a lower FLOP count, speeding up computations, and often pushing the layers' operations out of the FLOP-limited regime into the memory-limited regime.

We use the FLOP count for a representation layer as a proxy for measuring its runtime, because FLOPs dominate the runtime. We only consider the max of the multiply and add operations, since most modern processors have FMA or FMAC unit that can perform a simultaneous multiply-add operation. The typical way of applying representation layers to an input is through a matrix multiplication. Because of this, bias computations are ignored as the computation is dominated by the matrix multiplication.

The FLOP count is dominated by the duplication factor on the input blob size. A convolutional layer is a $4$D tensor in $\mathbb{R}^{c_o \times c_i \times k_h \times k_w}$, where $c_o$ is the number of output feature maps, $c_i$ is the number of input channels, and $k_h$ and $k_w$ are the kernel dimensions in $y$ and $x$ respectively. The convolution traverses an input blob in $\mathbb{R}^{h \times w}$ with stride $s_h$ in $y$ and $s_w$ in $x$. In a grouped convolution, the input and output channels are broken into $g$ groups and the $4$D weight tensor is in $\mathbb{R}^{c_o \times \frac{c_i}{g} \times k_h \times k_w}$. A grouped convolution is time-efficient because a batch process can do a group's worth of computation in a single step. For a deconvolution, the stride is applied on the output rather than the input. See Table \ref{tab:flopcounts} for FLOP counts for the forward pass of each representation layer.

\begin{table}[ht!]
\centering
\caption{Representation layers and FLOP counts}
\begin{tabular}{@{}ccc@{}} \toprule
\textbf{Convolution} & \textbf{Deconvolution} & \textbf{Fully-Connected} \\
\midrule
$\dfrac{hw c_i c_o k_h k_w}{s_h s_w g}$ & $\dfrac{hw c_i c_o k_h k_w s_w s_h}{g}$ & $hw c_i c_o$ \\
\bottomrule
\label{tab:flopcounts}
\end{tabular}
\end{table}

Other layers, such as batch normalization or activation layers, have a negligible number of FLOPs in the forward pass of a network, since only one operation per input pixel is typically applied.

\section{The Approximation Pipeline}

Only the model definition, the weights, and a single tunable input parameter $p$ that weights the importance of accuracy relative to speed on a scale from $0$ to $1$ are needed to apply DLA.

\begin{figure}[t]
  \centering
  \makebox[\linewidth][c]{\includegraphics[width=\linewidth]{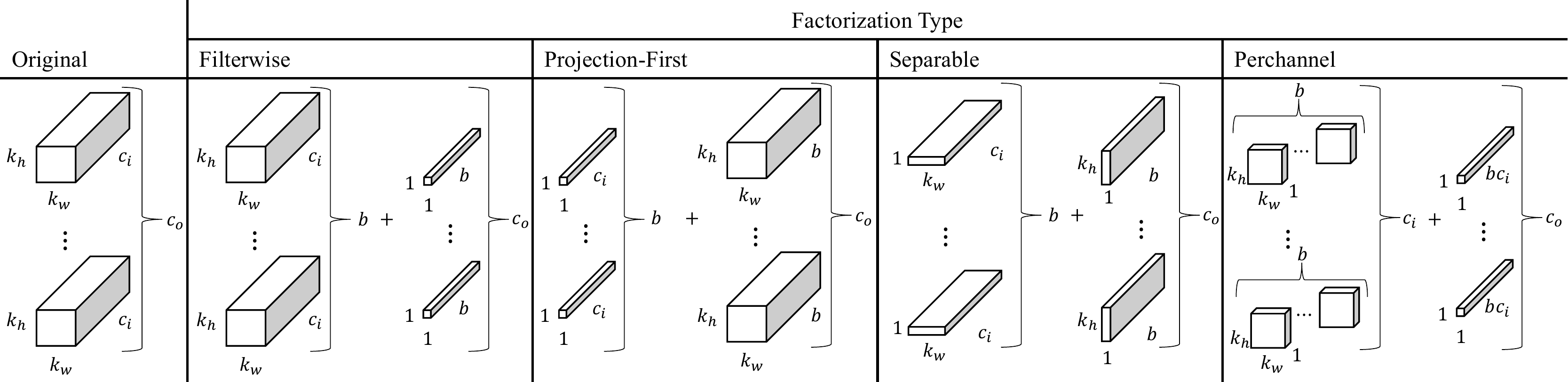}}
  \caption{A visual representation of the 4 factorization methods used in DLA. $c_i$ is the number of input channels, $c_o$ is the number of output channels, $k_w$ is the kernel width, $k_h$ is the kernel height, and $b$ is the factorization parameter.}
  \label{fig:graphic}
\end{figure}

The pipeline is applied sequentially to each representation layer in the network. There is no re-training done between steps, the entire process is done with the original trained network. Approximations are applied simultaneously to all layers at once. The discussion below is illustrated on convolutional layers, but similar extensions can be derived for fully-connected and deconvolutional layers. The pipeline is broken into two parts: (1) lossless optimizations and (2) lossy approximations. The lossless optimizations are applied first, since they have no accuracy impact and will always decrease runtime. The lossy approximations are based on low-rank approximations of representation layers using SVD \cite{jaderberg14speeding}, represented in Figure \ref{fig:graphic}. The optimal lossy approximation is chosen based on the accurary weighting parameter $p$.

\subsection{Lossless optimizations}

Here, the term \textit{lossless optimizations} means optimizations applied to the network that have no effect on accuracy. The main lossless optimization is the merging of adjacent linear layers. For example, representation layers are often followed by batch normalization and scale layers, to reduce overfitting during training time \cite{43442}. However, at deploy time, these operations are sequential linear operations applied on an input blob and can be combined into one. Table \ref{tab:linearops} shows how a cascade of linear layers, with $x$ and the input and $y$ as the output can be combined into a single representation layer.

\begin{table}[ht!]
\centering
\caption{Linear operators in neural networks}
\begin{tabular}{@{}llll@{}} \toprule
\textbf{Layer} & \textbf{Parameters} & \textbf{Operation} & \textbf{Cascaded operation} \\
\midrule
Representation & 
    \begin{tabular}{@{}ll@{}}$W$ & weight vector \\
                              $b$ & bias vector \\
                              & (per-channel) \\ \end{tabular}
    & $y_0 = Wx_0 + b$ & $y_0 = Wx_0 + b$ \\ \midrule
Batchnorm & 
    \begin{tabular}{@{}ll@{}}$\mu$ & learned mean \\
                                  & (per-channel) \\
                              $\sigma^2$ & learned variance \\
                                        & (per-channel) \end{tabular} & 
        $y_1 = \dfrac{x_1 - \mu}{\sigma}$ & $y_1 = \dfrac{Wx_0 + b - \mu}{\sigma}$ \\ \midrule
Scale & 
    \begin{tabular}{@{}ll@{}}$\alpha$ & learned constant \\
                                      & (per-channel) \\
                              $\beta$ & learned constant \\
                                      & (per-channel)  \end{tabular} &
          $y_2 = \alpha x_2 + \beta$ & $y_2 = \alpha \dfrac{Wx_0 + b - \mu}{\sigma} + \beta$ \\
\bottomrule
\label{tab:linearops}
\end{tabular}
\end{table}

The final cascaded operation, $y = \alpha \frac{Wx + b - \mu}{\sigma} + \beta$ has the same form as a representation layer and can therefore be performed with a single multiply-add operation, with the new weight parameter $W' = \alpha \frac{W}{\sigma}$ and new bias parameter $b' = \alpha \frac{b - \mu}{\sigma} + \beta$. Therefore, all three layers can be combined into a single representation layer with these new parameters. A similar derivation can apply for batchnorm / scale layers that appear before representation layers, if only a batchnorm layer (without a scale layer) is present, or if two representation layers are applied sequentially without a non-linearity in between.

While applying this lossless optimization saves a small number of FLOPs in the entire forward pass, it reduces the need for the neural network library to allocate memory to the batchnorm and scale layers, which reduces runtime. Because this optimization is exact (up to floating-point error), it should always be applied to a network before moving it to production.

\subsection{Lossy approximations}

After the lossless optimizations are applied to a layer, we then apply a series of lossy approximations. Each lossy approximation is a factorization based on SVD along a given axis of the representation layer. An \textbf{accuracy score} $A$ and \textbf{runtime score} $R$ is computed for each approximation. The overall score $S = p A + (1 - p) R$ is the weighted sum of the two with the user-specified parameter $p$. $A$ is the percentage of variation explained by an SVD approximation, and $R$ is the FLOP reduction. Intuitively, using low-rank decomposition to determine which FLOPs to keep and which to approximate means that representation layers that are redundant will have a lower-rank decomposition, and yield more speedup for less accuracy loss when approximated.

\subsubsection{Filter-wise factorization}

Filter-wise factorization applies SVD on the output-channels axis of the representation layer. The decomposition will be illustrated on the convolutional layer but is extensible for deconvolution and fully-connected layers without loss of generality. Applying SVD along the first axis of a convolution $W \in \mathbb{R}^{c_o \times c_i \times k_h \times k_w}$ yields a factorization $W = U \Sigma V^T$ with $s = \min(c_o, c_i k_h k_w)$ singular values, $U \in \mathbb{R}^{c_o \times s}$, $\Sigma$ is a diagonal matrix, and $V^T \in \mathbb{R} ^{s \times c_i \times k_h \times k_w}$. Taking the first $b$ singular values, we can reconstruct a low-rank approximation of $W = U^{\prime} \Sigma^{\prime} {V^T}^{\prime}$, where the inner dimension is changed from $s$ to $b$.

This can be expressed as two adjacent convolutions $W_1 = {V^T}^{\prime} \in \mathbb{R}^{b \times c_i \times k_h \times k_w}$ and $W_2 = U^{\prime} \Sigma^{\prime} \in \mathbb{R}^{c_o \times b \times 1 \times 1}$. Table \ref{tab:filterwiseflops} shows a detailed breakdown of FLOP counts and weight shapes for the resulting convolutions $W_1$ and $W_2$. There is a FLOP reduction in all cases where $b < \dfrac{c_i c_o k_w k_h}{c_i k_w k_h + g c_o}$. The ratio of original FLOPs to new FLOPs is the \textbf{runtime score} $R$ for this approximation. If $s_b$ is the $b$-th singular value of $W$, then the \textbf{accuracy score} $A$ is the percentage of variation explained by the first $b$ singular values $\dfrac{{s_b}^2}{\sum_{j = 0}^{c_o} {s_j}^2}$.

\begin{table}[ht!]
\centering
\caption{FLOP reduction for filter-wise factorization if $b$ singular values are used}
\begin{tabular}{@{}lccc@{}} \toprule
& \textbf{Original} & \multicolumn{2}{c}{\textbf{\qquad Factored}} \\\midrule
Names & $W$ & ${V^T}^{\prime}$ &  $U^{\prime} \Sigma^{\prime}$ \\
\addlinespace
Output channels & $c_o$ & $b$ &  $c_o$ \\
\addlinespace
Input channels & $c_i$ & $c_i$ &  $b$ \\
\addlinespace
Kernel size & $k_w \times k_h$ & $k_w \times k_h$ &  $1 \times 1$ \\
\addlinespace
FLOPs & $\dfrac{hw c_i c_o k_h k_w}{s_h s_w g}$ & $\dfrac{hw c_i b k_h k_w}{s_h s_w g}$ &  $\dfrac{hw b c_o}{s_h s_w}$ \\
\addlinespace
\bottomrule
\label{tab:filterwiseflops}
\end{tabular}
\end{table}

A variant of this approximation, called \textbf{projection-first factorization}, is applied the same way, except the $k_w \times k_h$ filter is reconstructed into the second convolution rather than the first. A similar calculation can be done for the FLOP reduction.

\subsubsection{Separable factorization}

Separable factorization applies SVD to the kernel axis of the representation layer. To find a separable factorization, the weight tensor $W$ is flattened into a matrix of size $c_o k_h \times c_i k_w$ and SVD is applied along the kernel axes. The resultant approximation tensors $W_1 \in \mathbb{R}^{b \times c_i \times k_h \times 1}$ and $W_2 \in \mathbb{R}^{c_o \times b \times 1 \times k_w}$ each have kernels that are $1$-dimensional in the spatial axes. Table \ref{tab:separableops} shows the detailed parameters to reconstruct the two separable weight tensors.

\begin{table}[ht!]
\centering
\caption{FLOP reduction for separable factorization if $b$ singular values are used}
\begin{tabular}{@{}lccc@{}} \toprule
& \textbf{Original} & \multicolumn{2}{c}{\textbf{\qquad Factored}} \\\midrule
Names & $W$ & ${V^T}^{\prime}$ &  $U^{\prime} \Sigma^{\prime}$ \\
\addlinespace
Output channels & $c_o$ & $b$ &  $c_o$ \\
\addlinespace
Input channels & $c_i$ & $c_i$ &  $b$ \\
\addlinespace
Kernel size & $k_w \times k_h$ & $k_w \times 1$ &  $1 \times k_h$ \\
\addlinespace
FLOPs & $\dfrac{hw c_i c_o k_h k_w}{s_h s_w g}$ & $\dfrac{hw c_i b k_h}{s_h s_w g}$ &  $\dfrac{hw b c_o k_w}{s_h s_w}$ \\
\addlinespace
\bottomrule
\label{tab:separableops}
\end{tabular}
\end{table}

Just like in the filter-wise factorization approximation, if $s_b$ is the $b$-th singular value of $W$, then the \textbf{accuracy score} $A$ is the percentage of the variation of $W$ explained by the first $b$ singular values $\dfrac{{s_b}^2}{\sum_{j = 0}^{c_o} {s_j}^2}$ and the \textbf{runtime score} $R$ is the FLOP reduction ratio.

\subsubsection{Per-channel factorization}

Per-channel approximation applies SVD to each input channel separately, to approximate the original convolution by a set of 2-dimensional convolutions to apply per-channel. $W$ is approximated by two new convolutions $W_1 \in \mathbb{R}^{bc_o \times \frac{c_i}{c_o} \times k_h \times k_w}$ and $W_2 \in \mathbb{R}^{c_o \times bc_o \times 1 \times 1}$. $W_1$ is applied in grouped fashion to the $c_i$ input channels. The result is $b c_o$ output channels after $W_1$. Then $W_2$ reconstructs the orignal $c_o$ output channels. See Table \ref{tab:perchannelops} for details.

\begin{table}[ht!]
\centering
\caption{FLOP reduction for perchannel factorization if $b$ singular values are used}
\begin{tabular}{@{}lccc@{}} \toprule
& \textbf{Original} & \multicolumn{2}{c}{\textbf{\qquad Factored}} \\\midrule
Names & $W$ & ${V^T}^{\prime}$ &  $U^{\prime} \Sigma^{\prime}$ \\
\addlinespace
Output channels & $c_o$ & $b c_o$ &  $c_o$ \\
\addlinespace
Input channels & $c_i$ & $c_i$ &  $b c_o$ \\
\addlinespace
Kernel size & $k_w \times k_h$ & $k_w \times k_h$ &  $1 \times 1$ \\
\addlinespace
Group & $1$ & $c_o$ &  $1$ \\
\addlinespace
FLOPs & $\dfrac{hw c_i c_o k_h k_w}{s_h s_w}$ & $\dfrac{hw c_i b k_h k_w}{s_h s_w}$ &  $\dfrac{hw b {c_o}^2}{s_h s_w}$ \\
\addlinespace
\bottomrule
\label{tab:perchannelops}
\end{tabular}
\end{table}

The decomposition is found by applying SVD $c_i$ times on the original weight $W_{c_o} \in \mathbb{R}^{c_o \times k_w k_h}$ that corresponds to the input channel, and the accuracy score $A$ is the percentage of variation of $W$ explained for that $b$ is averaged across all the input channels. Because the FLOP count is inversely proportional to the group parameter, this approximation results in a large FLOP reduction.

\subsection{Chaining approximations}

Approximations can be applied on top of one another. For example, if a projection-first approximation is applied on top of a filter-wise approximation on an input layer $W$, the resulting output will be three convolutions: $W_1 \in \mathbb{R}^{b_1 \times c_i \times 1 \times 1}$, $W_2 \in \mathbb{R}^{b_2 \times b_1 \times k_h \times k_w}$, and $W_3 \in \mathbb{R}^{c_o \times b_2 \times 1 \times 1}$. $R$ for this approximation chain is the ratio of the new FLOPs to the FLOPs from the original layer. $A$ is the product of the accuracy scores for each constituent decomposition. Since the convolutions are applied one after another, and any error introduced by the first will be carried over to the next.

\subsection{Optimizing the overall approximation}

The parameter $p$ a user specifies the relative importance of maintaining accuracy or decreasing runtime. The optimal approximation is the approximation (of all possible sequences of approximations) that has the highest score $S$. It is guaranteed to the best tradeoff between runtime and accuracy loss, since SVD is the optimal low-rank decomposition along the given axis.

\subsection{Finding approximation groups}

Some network architectures like ResNets have more complex structures, where layers in a network share the same input or the same output. In the case of a residual unit in a ResNet, pairs of layers share an output representation, since their outputs get added together as part of the residual operation. In these cases, the joined layers can be considered for approximation together as part of an approximation group. The $N$ weights are concatenated into a single weight matrix $W_c$, and the reconstructed weights after approximation are split into $N$ separate layer outputs to match the original $N$ output blobs.

\subsection{Applying approximations holistically}

During the forward pass of a network, errors introduced at layers closer to the input of the network propagate up, accumulating as the execution gets closer to the output and therefore are more consequential. Additionally, layers closer to the output tend to take a longer time to execute on a forward pass. To prevent high error accumulations, the parameter used for that layer's approximation is based on the user-defined parameter $p$, starting at $0.99$ and linearly decreasing to $p$ as you get farther from the input. The start value of $0.99$ can be further tuned to decrease the overall error. This also means that layers closer to the output (which typically are slower) have a more aggressive approximation applied, which provides a better trade-off between runtime and accuracy. Furthermore, in practice, $p$ is used as a threshold: no approximation whose accuracy score $A < p$ is considered a valid approximation.

\section{Experimental results}

On 4 network architectures designed specifically for computer vision applications, we show between 1.5x and 2x runtime improvement for a forward pass, between $5$ and $10\%$ loss in accuracy, and nearly full recovery of accuracy after finetuning. Runtime was tested using Caffe \citep{jia2014caffe} compiled with CUDA8 on a g2 cloud instance. Detailed accuracy and runtime results are shown in Table \ref{tab:results_acc}. The FLOP reduction (rather than memory reduction) correlates with the absolute speedup, with the exception of the ResNet50 network, as shown in Table \ref{tab:results_mults}. The ResNet50 network has many $1 \times 1$ convolutions that are by design memory-limited rather than FLOP-limited, so DLA is less effective.

\begin{table}[t!]
\centering
\caption{DLA applied to 4 networks and datasets}
\makebox[\linewidth][c]{
\begin{tabular}{@{}lllllll@{}} \toprule
& \multicolumn{2}{c}{\textbf{Runtime (ms)}} & & \multicolumn{3}{c}{\textbf{Accuracy (top-1 or mAP)}} \\
\addlinespace
\textbf{Network / dataset} & \textbf{Baseline} & \textbf{DLA} & \textbf{Speedup} & \textbf{Baseline} & \textbf{DLA} & \textbf{Finetuned} \\\midrule
AlexNet \citep{NIPS2012_4824} / CIFAR10 \citep{cifar10} & $6.06$ & $3.47$ & $1.75$x & $70.3\%$  & $67.75\%$  & $68.8\%$  \\
ResNet50 \citep{He2015} / ImageNet2012 \citep{Russakovsky:2015:ILS:2846547.2846559} & $40.9$ & $26.8$ & $1.50$x & $72.3\%$  & $62.2\%$  & $68.6\%$  \\
VGG16 \citep{Simonyan14c} / ImageNet2012 & $78.6$ & $45.2$ & $1.75$x & $70.38\%$  & $59.60\%$  & $70.4\%$  \\
YOLO \citep{redmon2016yolo9000} / VOC2007 \citep{pascal-voc-2007} & $63$ & $31$ & $2.00$x & $66.9$ & $61.9$ & $65.9$ \\
\bottomrule
\label{tab:results_acc}
\end{tabular}}
\end{table}

\begin{table}[t!]
\centering
\caption{Speedup, FLOP reduction, and memory reduction}
\makebox[\linewidth][c]{
\begin{tabular}{@{}llll@{}} \toprule
\textbf{Network / dataset} & \textbf{Speedup} & \textbf{FLOP Reduction} & \textbf{Memory Reduction} \\\midrule
AlexNet \citep{NIPS2012_4824} / CIFAR10 \citep{cifar10} & $1.75$x & $2.50$x & $1.10$x \\
ResNet50 \citep{He2015} / ImageNet2012 \citep{Russakovsky:2015:ILS:2846547.2846559} & $1.50$x & $1.20$x & $1.50$x \\
VGG16 \citep{Simonyan14c} / ImageNet2012 & $1.75$x & $1.70$x & $1.50$x \\
YOLO \citep{redmon2016yolo9000} / VOC2007 \citep{pascal-voc-2007} & $2.00$x & $2.00$x & $1.60$x \\
\bottomrule
\label{tab:results_mults}
\end{tabular}}
\end{table}

Varying the input parameter $p$ yields different points on an accuracy / runtime curve. This means that practitioners can choose what balance of accuracy and runtime is most important for the application. In Table \ref{tab:yolo}, decreasing values of $p$ yield faster runtimes (as measured in milliseconds on a g2 instance) but also higher accuracy losses. If time can be spent on fine-tuning the post-DLA output network, then accuracy can be recovered back to original levels.

\begin{table}[ht!]
\centering
\caption{Runtime and accuracy when varying $p$ on YOLO}
\begin{tabular}{@{}lll@{}} \toprule
\addlinespace
\textbf{$p$} & \textbf{Runtime (ms)} & \textbf{Accuracy (mAP)} \\\midrule
baseline & $63$ & $66.9$ \\
$0.9$ & $57$ & $66.9$ \\
$0.8$ & $54$ & $66.0$ \\
$0.7$ & $45$ & $65.4$ \\
$0.6$ & $41$ & $65.0$ \\
$0.5$ & $32$ & $61.9$ \\
$0.4$ & $27$ & $50.0$ \\
\bottomrule
\label{tab:yolo}
\end{tabular}
\end{table}

\section{Discussion}

Here we examine specific characteristics of DLA.

\paragraph{Memory} The optimization function for choosing which convolution to apply specifically targets FLOP reduction, since FLOP reduction is directly related to runtime decrease for FLOP-limited layers. However, a consequence of decreasing FLOP count by using fewer output channels in a cascade of convolutions is that less memory is needed to achieve a forward pass as well. Especially in embedded environments, this makes larger networks viable. After networks have been passed through DLA and see diminishing returns with FLOP-reduction methods, it means that memory movement is the bottleneck in runtime, and different memory-reducing optimizations should be applied. In Table \ref{tab:results_mults} we see memory requirements at runtime decrease as a result of applying DLA.

\paragraph{Upper bound} Because DLA applies FLOP-reducing approximations, any convolutions that are memory-limited will not be sped up significantly with DLA. For example, fully-connected layers and $1 \times 1$ convolutions are typically memory-limited. From Table \ref{tab:results_mults} we see that the ResNet50 network is more memory-limited than FLOP-limited, since the runtime speedup is more closely related to the memory rather than the FLOP reduction. The FLOP decrease is observed in the non-$1 \times 1$ convolutions, and the speedup is likely a result of the memory decrease rather than the FLOP decrease. Pushing beyond the $2$x speedup observed on YOLO without significant accuracy loss is not possible with the proposed DLA approximations, because once DLA has been applied, layers are moved from the FLOP-limited regime to the memory-limited regime in the roofline model.

\paragraph{GPU-specific optimizations} The runtime improvements here are reported on a g2 cloud instance, but proportional speedup is also observed on a CPU. For GPU targets, we have observed that although the relationship between FLOPs and runtime is linear overall, the single most significant factor in runtime for a given representation layer is the number of output channels. The runtime is a step function, with large jumps at output channels that are powers of $2$, and linear in between. So, in practice when targeting networks for GPUs, we only choose between decompositions with powers-of-2 output channels. For a CPU, this effect is not observed and any number of output channels are considered. GPUs have different FLOP and memory throughputs, which is are properties inherent to the GPU. This means absolute and relative speedups will be different according to the target GPU architecture. A benefit of DLA is that it is GPU-architecture-agnostic, meaning the FLOP count will always be decreased, which will typically result in some speedup on any target device.

\paragraph{Frameworks} Results were tested using both Caffe and MXNet frameworks, and DLA optimizations can be applied to either kind of model. Similar speedups are observed with both frameworks. This is because the FLOP reduction optimizations from DLA are model-agnostic: they operate on the weight matrices directly and are not dependent on software implementation. Exact observed runtime speedups will be different depending on framework and implementation, but speedups will be seen using all network formats.

\paragraph{Relationship to Training} Because we have shown that accuracy can be recovered by fine-tuning after applying DLA, a good strategy is to train large networks, then iteratively apply DLA and finetune, if the data and training procedure is available. More aggressive speedup can be achieved in this way, though will also take more time. On the other hand, DLA does not rely on data availability, and can be taken as a data-agnostic black box. Those who do not have access to the training data or procedure can just apply DLA to get speedup with minimal accuracy loss.

\section{Conclusion}

Deep Learning Approximation can be applied to an already-trained network to speed it up and incur only a small amount of accuracy loss. Access to training data is not required and the techniques are framework-agnostic, which means DLA can be used on black-box networks or in environments where the training data cannot be accessed. The combination of approximation that best achieves the desired accuracy loss is chosen for each layer through an exhaustive search. DLA can be combined with other methods for speeding up neural networks. This runtime reduction can generate a multiplier in cost reduction for a production service that uses neural networks. Any accuracy loss that that was introduced can be recovered by fine-tuning or re-training the new resultant network

\medskip

\small

\bibliographystyle{plainnat}
\bibliography{biblio}

\end{document}